\documentclass{article} 
\usepackage{iclr2027_conference,times}

\usepackage{amsmath,amsfonts,bm}

\def\eqref#1{equation~\ref{#1}}

\def\1{\bm{1}}

\DeclareMathAlphabet{\mathsfit}{\encodingdefault}{\sfdefault}{m}{sl}
\SetMathAlphabet{\mathsfit}{bold}{\encodingdefault}{\sfdefault}{bx}{n}

\usepackage{hyperref}
\usepackage{url}

\usepackage{amssymb}
\usepackage{graphicx}
\usepackage{booktabs}   
\usepackage{multirow}   
\usepackage{graphicx}   
\usepackage{xcolor}     
\usepackage{array}
\newlength{\datacolw}
\newcolumntype{C}{>{\centering\arraybackslash}p{\datacolw}}

\title{Anchors Enhance Controllability and Reasoning of Diffusion Language Models
}

\author{Feiyu Qu \\
Department of Computer Science \\
Dartmouth College \\
Department of Electrical and Computer Engineering \\
Carnegie Mellon University \\
\texttt{feiyu.qu.gr@dartmouth.edu} \\
\texttt{fqu@andrew.cmu.edu}
\And
Zheng Huang \\
Department of Computer Science \\
Dartmouth College \\
\texttt{zheng.huang.gr@dartmouth.edu}
\AND
Andrea Zanette \\
Department of Electrical and Computer Engineering \\
Carnegie Mellon University \\
\texttt{azanette@andrew.cmu.edu}
}

\iclrfinalcopy 
\begin{document}

\maketitle
\pagestyle{plain}
\thispagestyle{plain}
\begin{abstract}
Diffusion language models (DLMs) offer new opportunities for controllable text generation.
However, their controllability, defined as the ability to satisfy task constraints while preserving their original capabilities, remains underexplored.
DLMs’ bidirectional attention allows control tokens to guide generation from arbitrary positions in the output sequence, but effective guidance depends on the compatibility between their content and placement.
To this end, we propose AnchorDLM, a framework in which an anchor generator produces an anchor consisting of a short output fragment and its position for each query. 
The anchor is placed as control tokens at the specified position to initialize the output sequence of a frozen DLM and guide subsequent denoising.
We train the anchor generator with reinforcement learning (RL) to jointly optimize anchor content and placement, using rewards computed from the DLM's generated outputs based on answer correctness and constraint satisfaction.
Moreover, through the lens of discrete flow matching (DFM), we interpret anchor placement as helping to reshape the initial velocity field that guides generation.
Experiments on code generation and mathematical reasoning across five DLM backbones show that AnchorDLM improves controllability and further yields gains in reasoning performance across diverse benchmarks.

\end{abstract}

\section{Introduction}
\label{sec:intro}

Many generation tasks in real world demand more than fluency or correctness: they also require the ability to guide generation to satisfy the constraints of a given task, such as specific text structure~\citep{zhang2025style2code, jiang2024forward} or style~\citep{yang2025position, caliskan2015deanonymizing}.
We refer to the ability to satisfy such constraints while preserving the model's original capabilities as controllability~\citep{zhang2025style2code, liang2024controllable}. 
Compared with autoregressive (AR) language models, diffusion language models (DLMs)~\citep{austin2021structured, li2022diffusion, gong2022diffuseq, sahoo2024simple, li2025diffusion,huang2026diffusion} offer greater flexibility for controllable text generation.
DLMs generate text through iterative denoising with bidirectional attention, allowing each position to incorporate available context from both directions.
This bidirectional generation process allows control signals to be placed at arbitrary positions within the output sequence and to guide the generation of surrounding text.
However, such potential remains largely underexplored.

One way to realize this potential is to initialize the output sequence by placing control tokens\citep{xiong2025unveiling,lee2025unlocking}, which involves two intertwined decisions: what to provide and where to control.
Recent methods such as $S^3$~\citep{xiong2025unveiling} and Ti~\citep{lee2025unlocking} address these decisions by initializing generation with predefined structural templates containing elements such as JSON keys and delimiters.
Although effective for rigid format control, these methods rely on generic structural elements to determine what to provide, offering limited semantic guidance about the desired content or reasoning process.
For where to control, these methods rely on manually specified templates to determine token positions, limiting their ability to adapt placement to the control content and the needs of individual queries.
Addressing these limitations calls for adapting both control content and placement to each query while ensuring their compatibility, as control tokens become part of the generated output.

To improve controllability by learning what guidance to provide and where to place it, we propose AnchorDLM, a framework that trains an anchor generator to produce an anchor consisting of a short output fragment as control tokens and its placement position.
We learn the anchor content and its placement together so that each anchor provides semantic guidance suited to its position and can then improve constraint satisfaction without compromising task performance.
For each input query, we place the generated anchor at the specified position in the DLM's initial state to guide subsequent iterative denoising.
We optimize the anchor generator with reinforcement learning (RL), using rewards computed from the DLM's generated outputs based on answer correctness and constraint satisfaction.
We further offer a theoretical interpretation of anchor placement through the lens of discrete flow matching (DFM), in which anchors can be viewed as reshaping the initial velocity field that guides generation.
Our contributions are as follows:
\begin{enumerate}
\item We are the first to address controllability in DLMs through learnable anchors that provide semantic guidance. We further interpret the role of anchors through the lens of discrete flow matching.
\item We propose AnchorDLM, a model-agnostic framework comprising an anchor generator and a frozen DLM, in which the generator is trained in two stages to produce anchors that guide denoising from the DLM's initial state.
\item Extensive experiments across five DLM backbones show that AnchorDLM improves controllability by 11.30 G-mean points on average over the strongest same-backbone baselines and, when trained for reasoning, matches or exceeds them in all 20 reasoning settings.

\end{enumerate}
\begin{figure}[t]
\centering
\includegraphics[width=\linewidth]{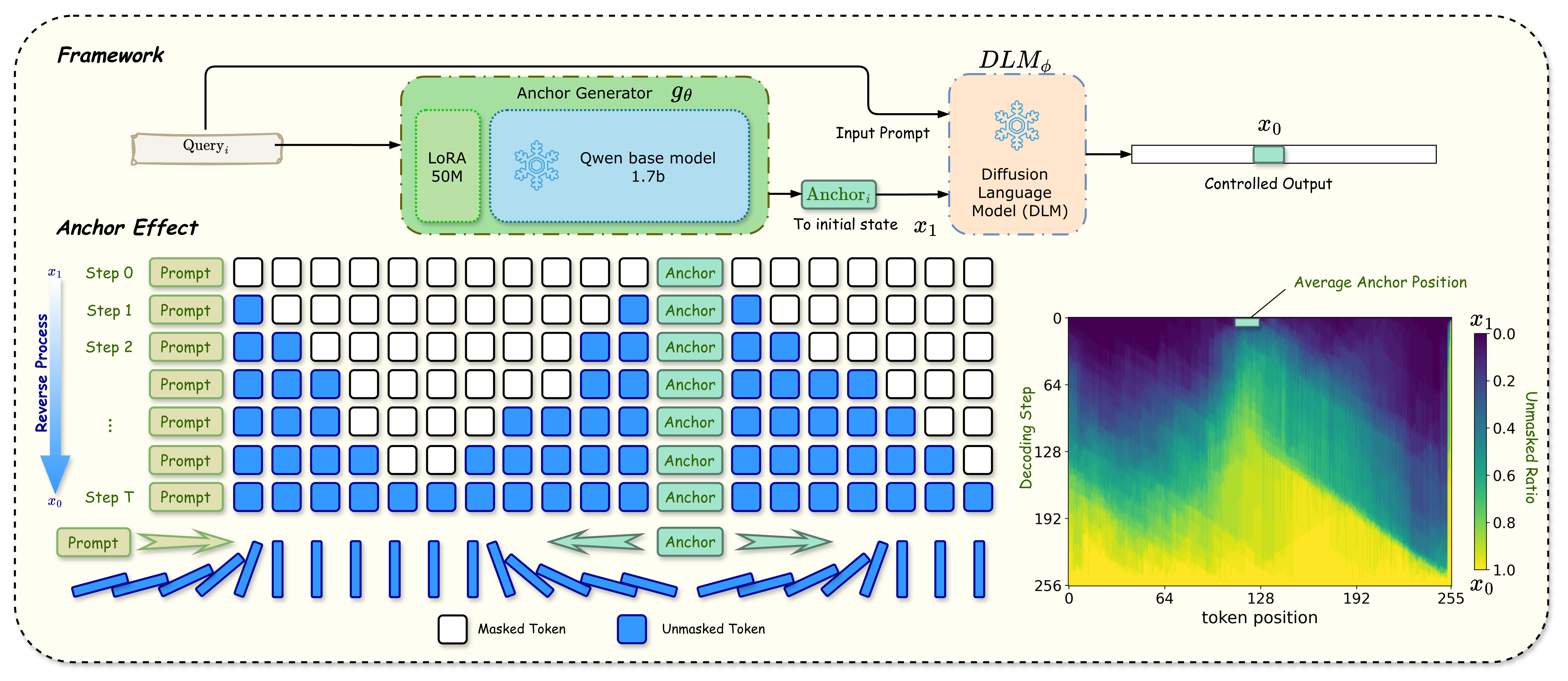}
\caption{Overview of AnchorDLM framework. (Right) Decoding order averaged over 50 samples.
}
\label{fig:framework}
\end{figure}

\section{Preliminary}
\label{sec:preliminary}

\subsection{Diffusion Language Models}
\label{sec:prelim_dlm}
In contrast to autoregressive models, diffusion language
models~\citep{austin2021structured,nie2025large,ye2025dream} capture the
text distribution through forward and reverse processes,
inspired by their continuous counterparts in image generation~\citep{austin2021structured}. 
We adopt the time convention $t\in[0,1]$ and let $x_t$ denote the partially noised sequence at time $t$, with $x_0$ being the clean sequence and $x_1$ a fully masked sequence.
The forward process independently replaces each token in $x_0$ with a mask symbol
$\mathrm{M}$ with probability $t$, gradually corrupting the sequence
until it is fully masked. The reverse process learns a mask
predictor $p_\phi(\cdot \mid x_t)$ parameterized by $\phi$ that recovers data distribution by iteratively predicting masked tokens as $t$ moves from 1 to 0. Training minimizes a weighted
cross-entropy loss evaluated only at masked positions~\citep{nie2025large}:
\begin{equation}
\mathcal{L}(\phi) \triangleq -\mathbb{E}_{t,x_0,x_t}\!\left[\frac{1}{t}
\sum_{i=1}^{L}\mathbf{1}[x_t^i = \mathrm{M}]\,
\log p_\phi(x_0^i \mid x_t)\right],
\end{equation}
where $x_0$ is drawn from the training data, $L$ is the sequence length, the factor $1/t$ compensates for the varying masking ratio across time
steps, the indicator $\mathbf{1}[\cdot]$ restricts the loss to
masked tokens and $x_t^i$ denotes the $i$-th token of $x_t$. We refer to this family of models as diffusion language
models (DLMs) throughout this paper.

\subsection{Discrete Flow Matching}
\label{sec:prelim_dfm}
Discrete flow matching (DFM)~\citep{gat2024discrete,campbell2024generative} provides a more general framework for
generative modeling over discrete state spaces by directly
parameterizing transition rates between sequences. Let
$\kappa_t \in [0,1]$ be a monotonically decreasing schedule with
$\kappa_0 = 1$ and $\kappa_1 = 0$, representing the weight retained on
the clean data at time $t$, and let $p_{0\mid t}(x^i \mid x_t)$ denote
the posterior over the clean token at position $i$ given the partially
noised sequence $x_t$. Under DFM, the transition rate at position $i$ is
\begin{equation}
\nu_t^i(x^i, x_t)
= \frac{-\dot{\kappa}_t}{1-\kappa_t}
\Bigl[p_{0\mid t}(x^i \mid x_t) - \delta_{x_t^i}(x^i)\Bigr],
\end{equation}
which is non-negative on off-diagonal entries since
$\dot{\kappa}_t \le 0$. A small backward Euler step samples the
previous state from
\begin{equation}
X_{t-h}^{i}
\sim \mathrm{Categorical}\!\Bigl(
  \delta_{X_t^i}(\cdot) + h\,\nu_t^i(\cdot,\, X_t)
\Bigr).
\end{equation}
where 
$\dot{\kappa}_t = \mathrm{d}\kappa_t / \mathrm{d}t$ denotes the time derivative, 
$h > 0$ denotes the small backward Euler step size, 
$\delta_{x_t^i}(\cdot)$ is the Kronecker delta indicator, equal to $1$ when its argument equals $x_t^i$ and $0$ otherwise, 
$X_t \in \mathcal{V}^L$ denotes the random sequence at time $t$ and $X_t^i$ its $i$-th token, with $\mathcal{V}$ the vocabulary augmented by the mask symbol $\mathrm{M}$.
This formulation highlights that DFM evolves probability mass through
context-dependent stochastic transition rates over a discrete state space. We will show in Section~\ref{sec:theory} that masked DLMs are a special case of this framework.

\section{Methodology}
\label{sec:methodology}

Our framework modifies the DLM's initial state by placing learned anchors, guiding the subsequent denoising toward controlled output. 
In this section, we first present the overall framework (Section~\ref{sec:framework}). 
We then describe a supervised fine-tuning stage over an augmented training set (Section~\ref{sec:sft}), that teaches the anchor generator to produce format valid anchors adaptive to each input query.
Building on this, in Section~\ref{sec:rl} we construct an offline reinforcement learning pipeline that uses rewards computed from the DLM's generated outputs based on answer correctness and constraint satisfaction, aligning the generator with the downstream DLM. 
Finally, in Section~\ref{sec:theory}, we provide insight into the effectiveness of anchors through the lens of DFM.

\subsection{Framework}
\label{sec:framework}

Given an input query $q$ and a frozen DLM that generates an output sequence $x \in \mathcal{V}^{L}$, our goal is to enhance controllability of the DLM without any update to its parameters.
To this end, our framework first produces the guidance that is placed into the DLM's initial state ahead of its generation, controlling the subsequent denoising toward the output aligning to the guidance.
We refer to these signals as anchors, each a pair $a = (p, c)$, represented as \texttt{\{"Position": "Content"\}}, where the \texttt{Position} $p \in \{0, 1, \ldots, L\}$ is a position index in the output sequence and the \texttt{Content} $c$ is a multi-token sequence over $\mathcal{V}$ placed starting at that position. 
As illustrated in Figure ~\ref{fig:framework}, the placed anchor affects decoding order and propagates information outward during denoising, influencing both its left and right context.

Our framework consists of two components bridged by an anchor: an anchor generator $g_\theta$ with parameters $\theta$, implemented as a lightweight Low-Rank Adaptation(LoRA) adapter on top of a pretrained language model(Qwen3-1.7B-Base~\citep{yang2025qwen3}), and a frozen DLM with fixed parameters $\phi$. Concretely, given the query $q$, the anchor generator produces an anchor $a = (p, c) = g_\theta(q)$.
We place $c$ at position $p$ of the DLM's initial state, keeping all remaining positions as the mask token $\mathrm{M}$, yielding an anchored initial state $x_1^{(a)} \in \mathcal{V}^{L}$. 
The frozen DLM performs standard denoising starting from $x_1^{(a)}$ to produce $x_0$. 
The single-step denoising transition can be written as
\begin{equation}
    p_\phi(x_{t-h} \mid x_t, q), \quad x_1 = x_1^{(a)},
\end{equation}
where $p_\phi(x_{t-h} \mid x_t, q)$ denotes the single-step denoising transition of the frozen DLM, and the anchor influences the entire generation by determining the initial state $x_1^{(a)}$.

\begin{figure}[t]
\centering
\includegraphics[width=\linewidth]{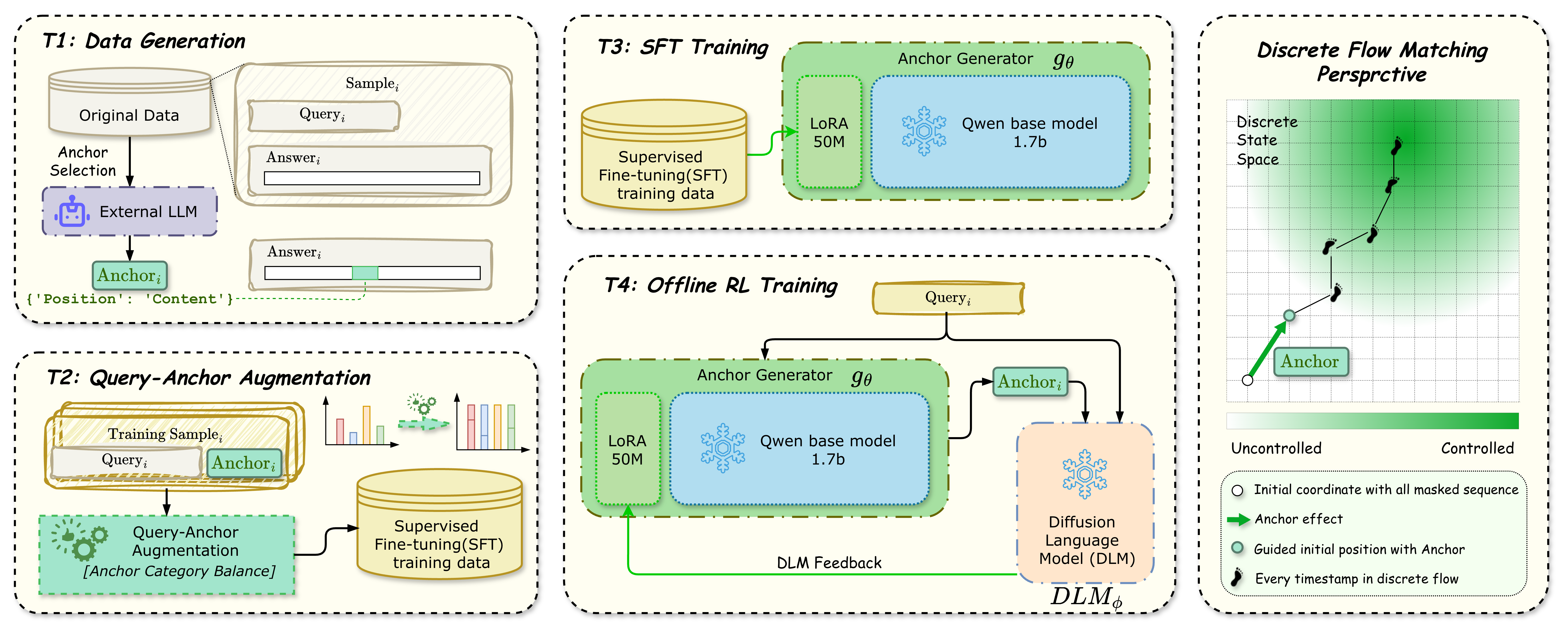}
\caption{Training pipeline and DFM perspective of AnchorDLM. T1--T2 construct training data. T3 performs supervised fine-tuning(SFT). T4 further refines the adapter with reinforcement learning(RL). (Right) From a discrete flow matching perspective, an anchor reshapes the initial coordinate of the discrete flow, pulling the generation trajectory from the uncontrolled region toward the controlled region of the discrete state space.}
\label{fig:pipeline_dfm}
\end{figure}

\subsection{Supervised Fine-tuning of the Anchor Generator}
\label{sec:sft}

This stage aims to equip the anchor generator with the ability to produce query adaptive and format valid anchors. 
To this end, we first use Claude Opus 4.6 to rewrite each ground truth answer into the target style, then extract a distinctive fragment and its position as the anchor target for the corresponding query (Figure~\ref{fig:pipeline_dfm}, T1).
To improve coverage of different anchor functions, we group anchors by functional role and rebalance the categories by paraphrasing queries associated with underrepresented categories (Query-Anchor Augmentation; Figure~\ref{fig:pipeline_dfm}, T2).

To teach the generator the required anchor format \texttt{\{"Position": "Content"\}}, we train it on the rebalanced dataset (See details in Appendix~\ref{appendix: imp} and~\ref{app:data} ).
However, anchors learned through SFT may still fail to guide the downstream DLM effectively, motivating further optimization.
\subsection{Reinforcement Learning of the Anchor Generator}
\label{sec:rl}
We further optimize the generator through reinforcement learning (RL), using rewards based on answer correctness and constraint satisfaction in the frozen DLM's outputs.
Given the cost of DLM inference, we use offline RL with a fixed preference dataset.
For each query, we sample $k$ anchors from the generator and evaluate the corresponding DLM outputs using these rewards.
Then we construct 80\% of preference pairs from queries with both successful and failed outputs, using the corresponding anchors as chosen and rejected responses, respectively.
For the remaining 20\%, we use queries where all $k$ outputs fail, pairing the supervised anchor as chosen with a sampled anchor as rejected.
This mixture combines preferences over sampled anchors with supervision for queries that yield no successful outputs~\citep{li2025simplemix, jiao2024preference, wang2025inco}.
To account for the mismatch between the current generator
and the data collection policy, we adopt Weighted Preference
Optimization (WPO)~\citep{zhou2024wpo}.
The WPO objective is:

\begin{equation}
\mathcal{L}_{\mathrm{WPO}} = -\mathbb{E}_{(q, a_w, a_l) \sim \mathcal{D}}\Bigl[
w_\theta(q, a_w) \, w_\theta(q, a_l) \, \log \sigma\Bigl(
\beta \log \frac{\pi_\theta(a_w \mid q)}{\pi_{\mathrm{ref}}(a_w \mid q)}
- \beta \log \frac{\pi_\theta(a_l \mid q)}{\pi_{\mathrm{ref}}(a_l \mid q)}
\Bigr)\Bigr],
\end{equation}
\begin{equation}
w_\theta(q, a) = \exp\!\left(\frac{1}{|a|}\sum_{n=1}^{|a|}
\left[\log \pi_\theta(a_n \mid q, a_{<n}) - \log \sum_{v \in \mathcal{V}} \pi_\theta(v \mid q, a_{<n})^2\right]\right),
\end{equation}
where $a_w$ and $a_l$ are the preferred and dispreferred anchors, $\pi_\theta$ is the current anchor generator and $\pi_{\mathrm{ref}}$ the SFT reference, $\beta$ controls the deviation from the reference, $|a|$ is the anchor's token length, and $a_n$ denotes the $n$-th token of $a$ (See details in Appendix~\ref{appendix: imp}). 
The weights $w_\theta(q, a)$ follow WPO's sampled alignment formulation and are detached from gradient computation~\citep{zhou2024wpo}.
Experiments show that this RL stage improves the control success rate of generated anchors on the downstream DLM.

\subsection{Interpretation from a Discrete Flow Matching Perspective}
\label{sec:theory}
We analyze AnchorDLM through the standard absorbing DFM with
independent masking, exact denoising posteriors, and sampling
in continuous time without remasking~\citep{gat2024discrete}.
For a fixed query $q$, let $p_0(\cdot\mid q)$ denote the
unanchored terminal distribution.
With $\kappa_t=1-t$, the probability velocity for reverse
updates from $t$ to $t-h$ is
\begin{equation}
\nu_t^i(v,x_t,q)
=
\frac{1}{t}
\left[
p_{0\mid t}^i(v\mid x_t,q)
-
\delta_{x_t^i}(v)
\right],
\quad 0<t<1,
\label{eq:dfm_velocity}
\end{equation}
where $p_{0\mid t}^i$ is the denoising posterior and $\delta$
denotes a point mass.
The masked prediction objective in Section~\ref{sec:prelim_dlm}
learns the posterior that parameterizes this velocity field.

For an anchor $a=(p,c)$, let $S_a$ denote the positions occupied
by $c$.
At revealed positions, the posterior is a point mass at the
observed token, so $\nu_t^i=0$.
Anchor tokens and subsequently revealed tokens therefore
persist, providing context that can influence velocity
evaluations throughout denoising
(Figure~\ref{fig:pipeline_dfm}).
For any correct controlled target $x^\star$ with
$x^\star_{S_a}=c$, initialization also reduces the Hamming
distance from $d_H(\mathrm{M}^{L},x^\star)=L$ to
$d_H(x_1^{(a)},x^\star)=L-|c|$.
Thus, a compatible anchor places the initial state closer
to a target while providing persistent guidance.

Define $E_a=\{x:x_{S_a}=c\}$ and assume
$p_0(E_a\mid q)>0$.
Fixing the anchor coordinates yields a conditional DFM on
the remaining positions, whose terminal law is
$p_0^{(a)}(\cdot\mid q)=p_0(\cdot\mid q,E_a)$.
Consequently,
\begin{equation}
p_0^{(a)}(x\mid q)
=
\frac{p_0(x\mid q)\,\mathbf{1}\{x\in E_a\}}{p_0(E_a\mid q)},
\qquad
D_{\mathrm{KL}}\!\left(p_0^{(a)}(\cdot\mid q)\,\Vert\,p_0(\cdot\mid q)\right)
=
-\log p_0(E_a\mid q).
\label{eq:anchor_terminal}
\end{equation}

The KL identity follows from the constant probability ratio
on $E_a$ and quantifies the information imposed by the anchor.

To connect this redistribution to controllability, let $C_q$
denote the event that an output is correct and satisfies
the task constraints, with $p_0(C_q\mid q)>0$.
Bayes' rule gives
\begin{equation}
p_0^{(a)}(C_q\mid q)
=
p_0(C_q\mid q)
\frac{p_0(E_a\mid C_q,q)}{p_0(E_a\mid q)}.
\label{eq:anchor_success}
\end{equation}
Thus, an anchor improves control success when it is more
likely among successful outputs than under the overall
output distribution.
This criterion motivates learning anchor content and
placement from the preferences constructed using downstream
outcomes in Section~\ref{sec:rl}.

\vspace{-2mm}

\section{Experiment}
\label{sec:experiment}
\vspace{-3mm}
This section begins with the experimental setup in Section~\ref{sec:setup}.
Section~\ref{sec:mainres} presents the improvements of AnchorDLM on DLMs' controllability and reasoning.
We then analyze the anchor generator's size (Section~\ref{sec:exp_size}), properties of the generated anchors including content and placement compatibility, diversity, and content ratio (Section~\ref{sec:anchor_analysis}), and the generator's generalization across DLMs (Section~\ref{sec:generalization}). 
We further conduct an ablation study on each component (Section~\ref{sec:ablation}).
Finally, Section~\ref{sec:case_study} provides a case study illustrating how anchors control DLM generation.

\subsection{Experiment Setup}
\label{sec:setup}

We evaluate AnchorDLM on mathematical reasoning benchmarks, GSM8K~\citep{cobbe2021training} and MATH500~\citep{lightman2024let}, and code generation benchmarks, HumanEval~\citep{chen2021evaluating} and MBPP~\citep{austin2021program}.
We maintain a strict separation between training and evaluation data throughout SFT and RL.
For GSM8K and MBPP, we use their official training splits.
For HumanEval, we construct training data by reformatting examples from the MBPP training split into the HumanEval task format.
For MATH500, we use samples from the MATH dataset after excluding all MATH500 evaluation problems.
All supervised anchors and RL preference pairs are constructed exclusively from these training sources.

We compare AnchorDLM against four categories of baselines:(i) the frozen DLM (LLaDA-8B-Base and LLaDA-8B-Ins~\citep{nie2025large}, Dream-7B-Base and Dream-7B-Ins \citep{ye2025dream}, and LLaDA 1.5-8B\citep{zhu2026llada} ); (ii) a prompt baseline that prepends a hand-crafted instruction to the input query; (iii) one recent controllable DLM method, $S^3$~\citep{xiong2025unveiling} , which places predefined schema templates into the DLM's initial state, and (iv)another  controllable DLM method, Ti~\citep{lee2025unlocking}, which places fixed reasoning templates with additionally adjusting template positions.

To evaluate controllability, we select four representative control objectives~\citep{yang2025position, jiang2024forward, zhang2025style2code,chen2023duetcs, buse2009learning, caliskan2015deanonymizing, hart2023eastwood} spanning content, structure, and reasoning logic. For each objective, we define a control score that measures the extent to which the generated output satisfies the specified requirement.

\textbf{Formal math(Fmt.)}~\citep{yang2025position}. The model is required to express solutions through mathematical formulas with minimal natural language. The control score is the proportion of formal mathematical expressions in the generated content.

\textbf{Backward reasoning(Bwd.)}~\citep{jiang2024forward}. 
The model is required to reason backward from the final goal to the premises needed to establish it.
The control score is the proportion of generated solutions that pass both an LLM assessment of backward reasoning and a check for characteristic reasoning connectives.

\textbf{Commented code(Cmt.)}~\citep{zhang2025style2code,buse2009learning, chen2023duetcs}. The model is required to produce code interleaved with comments to improve readability. The control score is the interleaving ratio between comments and code in the generated content.

\textbf{Structured code(Str.)}~\citep{caliskan2015deanonymizing,hart2023eastwood}.
The model is required to organize code into an initialization block followed by the main logic.
The control score is the proportion of generated solutions satisfying this abstract syntax tree(AST) based structural constraint.

To jointly consider whether control succeeds and whether the original task performance is preserved, following~\citet{mukherjee2025evaluating}and ~\citet{yang2023attractive}, we take the geometric mean(G-mean) of the control score and the corresponding task performance as the final composite controllability score, which is high only when control succeeds and task quality is preserved. Further details of the metrics are provided in the Appendix ~\ref{appendix: metric}.

\subsection{Main Results} 
\label{sec:mainres}
\vspace{-3mm}

Table~\ref{tab:main} reports results with anchor generators trained separately for controllability and reasoning.
AnchorDLM achieves the highest G-mean in all 30 controllability settings across five DLM backbones, exceeding the strongest baseline for the same backbone in each setting by 11.30 points on average.
The largest average gains occur on GSM8K: 22.41 points for formal mathematical expression and 21.01 points for backward reasoning.
When trained for reasoning, AnchorDLM matches or exceeds the strongest baseline for the same backbone in all 20 settings, with an average gain of 0.91 percentage points.
Compared with the prompted AR baseline Llama 3-8B-Base, AnchorDLM performs better in all 30 controllability settings and 18 of 20 reasoning settings.

\newcommand{\deltapos}[1]{{\fontsize{5pt}{6pt}\selectfont\textcolor{green!60!black}{(+#1)}}}
\newcommand{\deltaneg}[1]{{\fontsize{5pt}{6pt}\selectfont\textcolor{red!70!black}{(-#1)}}}
\newcommand{\dpv}[2]{#1\,\deltapos{#2}}
\newcommand{\dnv}[2]{#1\,\deltaneg{#2}}
\newcommand{\up}[1]{{\fontsize{6pt}{7pt}\selectfont\textcolor{green!60!black}{$\uparrow$#1}}}
\newcommand{\dn}[1]{{\fontsize{6pt}{7pt}\selectfont\textcolor{red!70!black}{$\downarrow$#1}}}
\begin{table}[t]
\vspace{-4mm}

\caption{Main results with the anchor generator trained for controllability (left) and reasoning (right). Controllability is measured by G-mean, and reasoning by accuracy or pass@1. Bold indicates the best result for each backbone.}
\label{tab:main}
\centering
\setlength{\tabcolsep}{2.5pt}
\renewcommand{\arraystretch}{1.1}
\resizebox{\linewidth}{!}{%
\begin{tabular}{ll|CCCCCC|CCCC}
\toprule
\multirow{3}{*}{Backbone} & \multirow{3}{*}{Method} & \multicolumn{6}{c|}{Controllability} & \multicolumn{4}{c}{Reasoning} \\
\cmidrule(lr){3-8} \cmidrule(lr){9-12}
 & & \multicolumn{2}{c}{GSM8K} & \multicolumn{2}{c}{HumanEval} & \multicolumn{2}{c|}{MBPP} & GSM8K & MATH & HE & MBPP \\
\cmidrule(lr){3-4} \cmidrule(lr){5-6} \cmidrule(lr){7-8} \cmidrule(lr){9-9} \cmidrule(lr){10-10} \cmidrule(lr){11-11} \cmidrule(lr){12-12}
 & & Fmt. & Bwd. & Cmt. & Str. & Cmt. & Str. & Acc & Acc & Pass@1 & Pass@1 \\
\midrule
\multirow{2}{*}{Llama 3-8B-Base}
 & Vanilla       & 13.92 & 10.02 & 23.49 & 44.13 & 25.82 & 45.65 & 43.59 & 24.80 & 33.54 & 41.40 \\
 & Prompt        & 24.66 & 21.67 & 36.73 & 46.97 & 39.61 & 47.41 & 56.03 & 28.00 & 36.59 & 43.80 \\

\midrule
\multirow{5}{*}{LLaDA-8B-Base}
 & Vanilla       & 5.01  & 5.01  & 14.59 & 42.32 & 29.77 & 43.16 & 66.03 & 30.20 & 34.15 & 39.80 \\
 & Prompt        & 7.99  & 7.40  & 23.29 & 45.04 & 31.16 & 47.27 & 66.57 & 31.80 & 35.37 & 40.20 \\
 & S$^{3}$       & 36.49 & 34.55 & 17.70 & 35.20 & 30.30 & 42.00 & 52.16 & 32.20 & 21.34 & 16.80 \\
 & Ti            & 38.58 & 35.19 & 16.86 & 45.43 & 30.11 & 46.81 & 49.89 & 31.60 & 28.05 & 17.80 \\
 & Ours$^{\dagger}$ & \textbf{69.33} & \textbf{71.08} & \textbf{43.89} & \textbf{50.41} & \textbf{40.05} & \textbf{49.27} & \textbf{67.17} & \textbf{33.40} & \textbf{37.20} & \textbf{43.60} \\
\midrule
\multirow{5}{*}{LLaDA-8B-Ins}
 & Vanilla       & 17.08 & 30.94 & 29.83 & 35.85 & 28.25 & 32.06 & 56.10 & 33.80 & 39.02 & 40.80 \\
 & Prompt        & 16.83 & 32.02 & 45.64 & 47.34 & 48.35 & 45.75 & 57.77 & 34.60 & 40.24 & 40.80 \\
 & S$^{3}$       & 45.33 & 49.15 & 38.77 & 44.46 & 33.11 & 38.15 & 52.16 & 33.20 & 36.59 & 25.00 \\
 & Ti            & 45.36 & 47.85 & 37.64 & 47.51 & 29.52 & 46.65 & 58.07 & 34.40 & 32.93 & 18.80 \\
 & Ours$^{\dagger}$ & \textbf{61.85} & \textbf{65.72} & \textbf{46.07} & \textbf{52.40} & \textbf{49.96} & \textbf{50.90} & \textbf{58.53} & \textbf{34.80} & \textbf{41.46} & \textbf{41.40} \\
 \midrule
\multirow{5}{*}{LLaDA 1.5-8B}
 & Vanilla              &18.65     & 31.27    &33.81     &41.02     &30.94     &41.14      & 59.14 & 33.40     & 39.63 & 44.00 \\
 & Prompt        &23.12   &34.28   &42.56   &46.69   &38.01   &47.76   &61.41   &34.20   &\textbf{40.24}   &   45.40\\
 & S$^{3}$       & 51.47  & 53.54  & 39.02  &  42.51 & 37.41  &41.63   &60.27   &33.60   &37.80   &   43.80\\
 & Ti            &46.69   & 50.17  & 39.37  &47.78   & 34.62  & 44.46  &60.80   &34.20   &35.98   &  44.80\\
 & Ours$^{\dagger}$     &\textbf{70.08}   &\textbf{67.42}   &\textbf{47.11}   &\textbf{54.53}   &\textbf{49.65}   &\textbf{52.48}      & \textbf{61.56} & \textbf{34.80}     & \textbf{40.24} & \textbf{47.40} \\
\midrule
\multirow{5}{*}{Dream-7B-Base}
 & Vanilla       & 17.52 & 15.24 & 18.36 & 45.27 & 31.62 & 42.77 & 73.69 & 37.00 & 52.44 & 50.80 \\
 & Prompt        & 25.35 & 23.09 & 31.01 & 47.92 & 38.75 & 48.27 & 74.83 & 40.20 & 54.88 & 52.60 \\
 & S$^{3}$       & 38.74 & 37.29 & 27.64 & 42.32 & 37.88 & 43.87 & 72.93 & 39.60 & 48.78 & 41.80 \\
 & Ti            & 43.63 & 41.77 & 26.15 & 49.06 & 35.41 & 45.59 & 71.04 & 40.60 & 51.83 & 45.00 \\
 & Ours$^{\dagger}$ & \textbf{72.61} & \textbf{71.70} & \textbf{46.79} & \textbf{55.12} & \textbf{44.05} & \textbf{53.24} & \textbf{75.36} & \textbf{41.80} & \textbf{56.10} & \textbf{53.20} \\
\midrule
\multirow{5}{*}{Dream-7B-Ins}
 & Vanilla       & 20.51 & 33.51 & 37.13 & 44.48 & 29.41 & 35.81 & 75.59 & 38.00 & 48.78 & 52.00 \\
 & Prompt        & 29.22 & 42.16 & 50.52 & 55.17 & 44.15 & 51.18 & 76.50 & 42.40 & 52.44 & 54.00 \\
 & S$^{3}$       & 49.05 & 51.13 & 46.63 & 49.27 & 39.59 & 47.72 & 74.22 & 41.80 & 53.66 & 48.80 \\
 & Ti            & 48.69 & 49.43 & 43.39 & 53.74 & 38.61 & 50.82 & 75.89 & 41.80 & 51.83 & 50.60 \\
 & Ours$^{\dagger}$ & \textbf{66.26} & \textbf{59.93} & \textbf{51.78} & \textbf{58.44} & \textbf{48.92} & \textbf{56.39} & \textbf{76.80} & \textbf{43.40} & \textbf{54.27} & \textbf{54.40} \\

\bottomrule
\end{tabular}%
}
\vspace{-5mm}
\end{table}

\subsection{Anchor Generator Size and Capacity}
\label{sec:exp_size}
\vspace{-3mm}

We compare the downstream performance of anchors extracted from models of different sizes on HumanEval and MBPP.
We use the same heuristic to extract anchors from each model's complete outputs.
Table~\ref{tab:size} shows clear gains from 0.6B to 1.7B, while further improvements vary across benchmarks and model sizes.
We adopt Qwen3-1.7B as the default $g_\theta$ to balance model size and downstream performance.

\begin{table}[t]
\centering
\caption{Pass@1 (\%) with anchors extracted from
different models using the same heuristic.}
\label{tab:size}
\small
\setlength{\tabcolsep}{5pt}
\begin{tabular}{@{}lrrrrrr@{}}
\toprule
Model & \multicolumn{5}{c}{Qwen3} & Llama 3.3 \\
\cmidrule(lr){2-6}\cmidrule(l){7-7}
Size & 0.6B & 1.7B & 4B & 8B & 32B & 70B \\
\midrule
HumanEval
& 24.39 &  33.54 &  38.41
&  38.41 &  35.98 &  40.24 \\
MBPP
&  28.60 &  37.80 &  37.60
&  38.00 &  42.20 &  43.20 \\
\bottomrule
\end{tabular}
\vspace{-5mm}
\end{table}

\vspace{-1mm}

\subsection{Anchor Analysis}
\label{sec:anchor_analysis}
To understand how anchors guide DLM generation, we examine
three aspects: (i) the compatibility between anchor content
and placement; (ii) the anchor position distribution; and (iii) the proportion of anchor content in the final generation.

\vspace{-3mm}

\paragraph{Content and placement compatibility.} 

\begin{figure}[t]
\centering
\includegraphics[width=\linewidth]{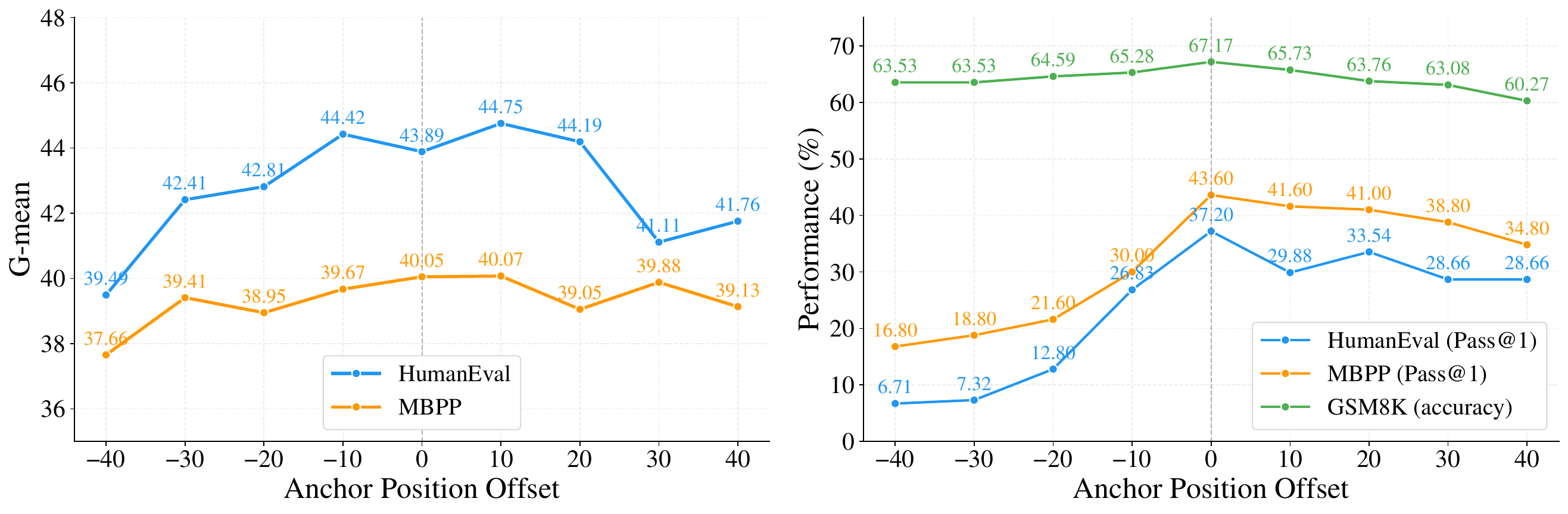}
\caption{Effect of anchor placement with content held fixed: G-mean (left) and performance (right).}
\label{fig:position}
\end{figure}

We analyze anchor placement, the proportion of anchor content in generated outputs, and generalization across inputs.
To examine the importance of matching anchor content and placement,
we use LLaDA-8B-Base as the frozen backbone, hold the anchor content
fixed, and shift its position by offsets ranging from $-40$ to $+40$ tokens.
Across these offsets, G-mean varies by 5.3 points on HumanEval and 2.4 points on MBPP (Figure~\ref{fig:position}).
Moving anchors earlier reduces code generation pass@1 by up to 30.5 percentage points, whereas later shifts cause smaller declines and GSM8K accuracy varies by 6.9 percentage points across all offsets.
These results show that relocating the same anchor content can impair task performance, highlighting the importance of matching content and placement, particularly for code generation.


\vspace{-3mm}

\begin{figure}[t]
\vspace{-2mm}

\centering
\includegraphics[width=0.9\linewidth]{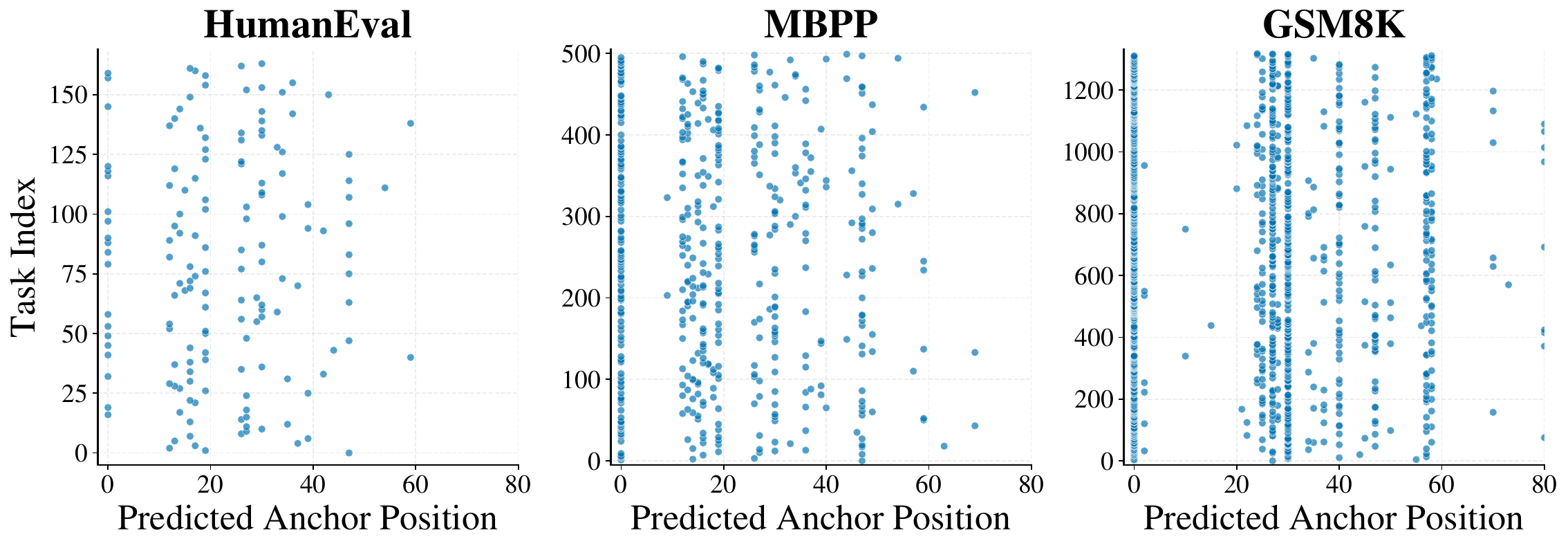}
\caption{Anchor position distribution and density across datasets.}
\label{fig:anchor_pos_dist}
\end{figure}

\paragraph{Anchor position distribution.} 
Figure~\ref{fig:anchor_pos_dist} visualizes the distribution and density of anchor positions predicted by our generator across datasets.
Since our method predicts anchor positions adaptively per query, the distribution is broad, with predicted positions spanning over 80 tokens across samples.
This observation confirms that our method offers substantial diversity and flexibility of anchors.

\vspace{-3mm}

\paragraph{Anchor content ratio.} 
To verify that AnchorDLM's gains do not come from placing large amounts of pre-computed content into the output, we measure the average fraction of output tokens contributed by the anchor on each benchmark.
Across benchmarks, anchors account for only 4.33--10.88\% of output tokens on average (Appendix~\ref{app:content_ratio}).


\vspace{-3mm}

\subsection{Generalization across DLMs}
\label{sec:generalization}

To examine whether the anchor generator generalizes across DLMs, we transfer $g_\theta$, trained using outputs from LLaDA-8B-Base, to LLaDA-8B-Instruct without retraining.
On the target DLM, it improves all evaluated metrics over the vanilla baseline, with controllability gains of 15.19 to 42.43 points (Appendix~\ref{app:generalization}).


\vspace{-2mm}

\subsection{Ablation Study}
\label{sec:ablation}
\vspace{-3mm}

We ablate three main design choices of AnchorDLM: removing the query-anchor augmentation, removing the RL stage (keeping only SFT), and replacing WPO with vanilla DPO. As shown in Table~\ref{tab:ablation}, all three components contribute positively to the final performance. 
Removing the query-anchor augmentation is by far the most damaging, nearly eliminating the gain of AnchorDLM and even hurting task performance, because the raw anchor distribution is heavily imbalanced across categories, leading generator overfitting to specific cases. Without RL, the generator cannot refine anchors using DLM feedback; replacing WPO with DPO introduces a distributional gap that makes the feedback signal stale.

\begin{table}[!htbp]
\caption{Ablation study of AnchorDLM on GSM8K and MBPP.}
\label{tab:ablation}
\centering
\small
\setlength{\tabcolsep}{3pt}
\resizebox{0.8\linewidth}{!}{
\begin{tabular}{l|cc|cc|c}
\toprule
\multirow{2}{*}{Variant} & \multicolumn{2}{c|}{GSM8K} & \multicolumn{2}{c|}{MBPP} & \multirow{2}{*}{Avg.\ $\Delta$} \\
& Formal & Acc & Comt. & Pass@1 & \\
\midrule
LLaDA-8B-Base (no anchor)        & 5.01 & 66.03 & 29.77 & 39.80 & -- \\
\midrule
\textbf{Full AnchorDLM}               & \textbf{69.33} \up{64.32} & \textbf{67.17} \up{1.14} & \textbf{40.05} \up{10.28} & \textbf{43.60} \up{3.80} & \textbf{+19.89} \\
\quad w/o WPO (replace with DPO)      & 60.42 \up{55.41} & 66.79 \up{0.76} & 38.06 \up{8.29} & 40.20 \up{0.40} & +16.22 \\
\quad w/o RL (SFT only)               & 47.16 \up{42.15} & 65.96 \dn{0.07} & 36.88 \up{7.11} & 40.00 \up{0.20} & +12.35 \\
\quad w/o query-anchor augmentation & 22.97 \up{17.96} & 58.98 \dn{7.05} & 31.73 \up{1.96} & 28.00 \dn{11.80} & +0.27 \\
\bottomrule
\end{tabular}
}
\end{table}

\subsection{Case Study}
\label{sec:case_study}
\vspace{-2mm}

\begin{figure}[t]
\centering
\includegraphics[width=0.95\linewidth]{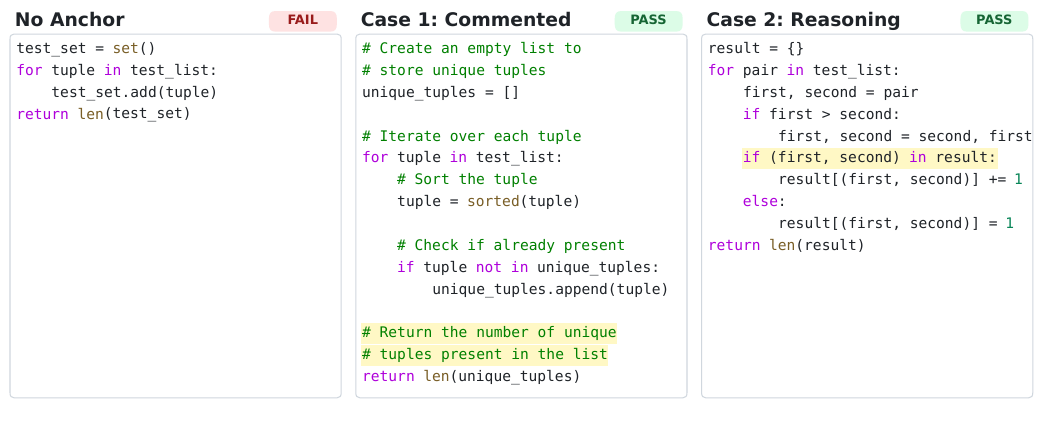}
\caption{Case study on MBPP sample (idx=389), with anchor highlighted. }
\label{fig:case_study}
\end{figure}

Figure~\ref{fig:case_study} illustrates anchor effects on MBPP task (idx=389), which requires counting distinct unordered pairs.
Without an anchor, the DLM incorrectly treats reversed pairs as distinct and fails the tests.
With a comment anchor (Case 1), the DLM generates a correct solution with interleaved comments and code, satisfying both correctness and style constraints.
With a reasoning anchor (Case 2), the DLM correctly orders each pair before dictionary lookup to count distinct pairs.

\vspace{-3mm}

\section{Related Works}
\label{sec:related_works}
\vspace{-2mm}

\paragraph{Diffusion language models.} Diffusion language models (DLMs)~\citep{austin2021structured, li2022diffusion, gong2022diffuseq, sahoo2024simple, shi2024simplified,lou2023discrete,khanna2025mercury} have recently emerged as a parallel, bidirectional alternative to autoregressive generation. Masked DLMs such as LLaDA~\citep{nie2025large, zhu2025llada} and Dream~\citep{ye2025dream} iteratively denoise a fully masked sequence under bidirectional attention. These models have shown competitive performance on language modeling, code generation, and mathematical reasoning.
\vspace{-3mm}

\paragraph{Controllable text generation.} Controlling generation has been extensively studied in the auto-regressive setting through prompting~\citep{wei2021finetuned}, classifier guidance~\citep{sitdikov2022classifiers}, and constrained decoding~\citep{geng2023grammar, mundler2025type, sanchez2023stay}. For DLMs, recent work has begun to explore control mechanisms tailored to bidirectional, masked decoding. $S^{3}$~\citep{xiong2025unveiling} injects predefined schematic templates (e.g., JSON keys, brackets, or fixed scaffolds) into the initial state to enforce structural formats, and Ti~\citep{lee2025unlocking} similarly relies on hand-designed structural markers with adjustable positions. While effective on rigid format control tasks, these approaches use fixed tokens, providing little semantic guidance that the DLM can utilize. In contrast, our framework learns both the position and the semantic content of the placed anchor end-to-end with feedback from the downstream DLM.

\vspace{-3mm}

\section{Conclusion}
\label{sec:conclusion}
\vspace{-2mm}


In this work, to address challenge of controllability of DLM, we introduced AnchorDLM, a model agnostic framework that jointly learns anchor content and placement for each query to guide generation from the initial state of a frozen DLM.
Our DFM analysis characterizes the persistent influence of anchors on the velocity field and their effect on the terminal distribution under idealized assumptions.
Across five DLM backbones, training for the respective objectives improves controllability by 11.30 G-mean points on average and matches or exceeds the strongest same-backbone baseline in every reasoning setting.

\medskip

\newpage
\bibliography{ref}
\bibliographystyle{iclr2027_conference}

\section{Appendix}


\subsection{Implementation Details}
\label{appendix: imp}
We instantiate the frozen DLM as LLaDA-8B-Base and LLaDA-8B-Instruct~\cite{nie2025large}, and the anchor generator $g_\theta$ as a Qwen3-1.7B backbone with a LoRA adapter of $50$M trainable parameters.

\paragraph{Supervised fine-tuning.}
We adopt a two-stage SFT schedule. Stage~1 trains a prefix anchor on SETUP-only data with $p^*\!=\!0$ for $5$ epochs at learning rate $2\!\times\!10^{-4}$, where $p^*$ denotes the supervised target position of the anchor and $p^*\!=\!0$ corresponds to placing the anchor as a prefix of the generation. Stage~2 continues training on the full rebalanced data for $3$ epochs at learning rate $1\!\times\!10^{-4}$. Both stages share the same LoRA configuration, batch size $4$ with gradient accumulation $2$, maximum sequence length $1536$, linear schedule with $10\%$ warmup, and \texttt{bf16} precision. 

\paragraph{Reinforcement learning.}
We use k=8 rollouts, with temperature = 1.0.
Starting from the SFT checkpoint, we further train with WPO for $4$ epochs at learning rate $1\!\times\!10^{-5}$ and $\beta\!=\!0.3$, using batch size $4$ with gradient accumulation $2$, maximum sequence length $1536$, linear schedule with $10\%$ warmup, and \texttt{bf16} precision.

\paragraph{Compute resources.}
All training and evaluation are conducted on NVIDIA L40 GPUs. The dominant cost is DLM inference during evaluation and during preference-data collection for RL: a single LLaDA-8B inference sample takes on average roughly $20$ seconds on coding benchmarks (HumanEval, MBPP) and roughly $30$ seconds on the mathematical reasoning benchmark (GSM8K, MATH500).

\subsection{Data Construction Details}
\label{app:data}

We provide additional details on anchor categorization and the Query-Anchor Augmentation introduced in Section~\ref{sec:sft}. We use a four-category scheme for both mathematical reasoning (GSM8K, MATH500) and code generation (HumanEval, MBPP), with the categories themselves defined separately for each task family.

\paragraph{Math: anchor categories.}
Each anchor is assigned to one of four categories based on its functional role in the reasoning chain:
\begin{itemize}
    \item \textbf{SETUP}: the set up reasoning step that interprets the question.
    \item \textbf{TRANSFORM}: an intermediate computation that converts existing values into a new value.
    \item \textbf{ACCUM}: an aggregation step that combines values, indicated by cues such as \emph{total}, \emph{altogether}, \emph{combined}.
    \item \textbf{DIFF}: a comparison or remainder step, indicated by cues such as \emph{more than}, \emph{less than}, \emph{remaining}, \emph{difference}.
\end{itemize}

\paragraph{Math: category rebalancing.}
The raw distribution is highly imbalanced: $\mathrm{SETUP}$ ($53.7\%$), $\mathrm{TRANSFORM}$ ($30.4\%$), $\mathrm{ACCUM}$ ($12.5\%$), $\mathrm{DIFF}$ ($3.4\%$). We rebalance all four categories to a uniform target of $889$ samples each (the original size of $\mathrm{ACCUM}$): $\mathrm{SETUP}$ and $\mathrm{TRANSFORM}$ are downsampled, $\mathrm{ACCUM}$ is kept, and $\mathrm{DIFF}$ is upsampled via Query-Anchor Augmentation. 

\paragraph{Math: Query-Anchor Augmentation.}
For the under-represented $\mathrm{DIFF}$ category, we generate paraphrased queries while keeping the original anchor and answer unchanged. The paraphrasing applies lightweight, semantics-preserving substitutions on question wording, including:
\emph{How many} $\leftrightarrow$ \emph{What is the total number of};
\emph{How much} $\leftrightarrow$ \emph{What is the total amount};
\emph{in total} $\leftrightarrow$ \emph{altogether};
\emph{remaining} $\leftrightarrow$ \emph{left over};
\emph{bought} $\leftrightarrow$ \emph{purchased};
\emph{needs to} $\leftrightarrow$ \emph{has to}.

\paragraph{Code: anchor categories.}
For code generation, anchors are categorized by their syntactic role in the program:
\begin{itemize}
    \item \textbf{COND}: a key decision point, e.g., an \texttt{if} or \texttt{while} condition.
    \item \textbf{EXPR}: the core computation expression, typically near the top of the function body.
    \item \textbf{BUILD}: a step that constructs the result, e.g., \texttt{append}, \texttt{add}, \texttt{extend}, \texttt{yield}.
    \item \textbf{UPDATE}: a state update inside a loop, e.g., accumulators, assignments, or DP transitions.
\end{itemize}

\paragraph{Code: Query-Anchor Augmentation.}
For code, we augment the query side by paraphrasing docstrings, varying parameter names, and shuffling doctest order, again without modifying the anchor or the reference solution. After augmentation, the four categories are balanced to the size of the largest category.

\paragraph{Anchor construction prompts.}
The supervised anchors used in Stage~1 and Stage~2 of SFT are obtained by prompting an external LLM (Claude Opus~4.6) to (i) rewrite each reference answer in the target style and (ii) extract the most distinctive sentence/line as the anchor. We use one prompt per (benchmark, control-objective) pair. All prompts share the same shape: \emph{system instruction} $\rightarrow$ \emph{style description} $\rightarrow$ \emph{rewriting and extraction rules}. We only show the style description and the rewriting/extraction rules below; the anchor format \texttt{\{"Position": "Content"\}} is appended to every prompt as the required output schema.

\subparagraph{GSM8K / MATH500 — backward reasoning.}
\begin{quote}\small
\textbf{Style.} Backward reasoning solves a problem by starting from the final goal and recursively decomposing it into the sub-goals or quantities required to obtain it, only resolving each sub-goal once its dependencies have been identified. Rewrite the solution in this goal-first, backward style: explicitly state the final goal and the quantities needed to obtain it before doing any computation, using phrases such as ``To find $X$, we need $Y$\,'' or ``Working backwards from the goal\,''. Do \emph{not} use forward connectives such as ``First / Then / Finally''.\\
\textbf{Rewrite rule.} Keep the final numerical answer unchanged and the math correct. Do not introduce information beyond the reference answer.\\
\textbf{Anchor extraction.} Pick the single most pivotal step in the backward decomposition (the one whose sub-goal most directly determines the rest of the chain) as the anchor.
\end{quote}

\subparagraph{GSM8K / MATH500 — formal math.}
\begin{quote}\small
\textbf{Style.} Rewrite the solution as a sequence of symbolic equations of the form \texttt{Let $x_n = \ldots$}, with minimal natural language. Each intermediate quantity is bound to a named variable and reused by subsequent lines.\\
\textbf{Rewrite rule.} Preserve the reference numerical answer. Do not skip computation steps and do not introduce variables that are never used.\\
\textbf{Anchor extraction.} Pick the single \texttt{Let $x_n = \ldots$} line that captures the central transformation of the problem as the anchor.
\end{quote}

\subparagraph{HumanEval / MBPP — commented code.}
\begin{quote}\small
\textbf{Style.} Rewrite the reference solution so that comments are interleaved with code throughout the function body, not dumped as a leading block. Each non-trivial code line should be preceded or followed by a short, meaningful comment that describes intent.\\
\textbf{Rewrite rule.} Keep the function signature, behavior, and test results identical to the reference solution. Do not insert decorative or redundant comments.\\
\textbf{Anchor extraction.} Pick the single \texttt{[code, comment]} pair that most clearly conveys the algorithmic intent of the solution as the anchor.
\end{quote}
\subparagraph{HumanEval / MBPP — structured code.}
\begin{quote}\small
\textbf{Style.} Rewrite the reference solution so that the function body begins with a contiguous variable initialization block, followed by the main logic. A function docstring and necessary early-exit guards may precede this block. Use literal values or empty built-in containers for initialization where appropriate.\\
\textbf{Rewrite rule.} Keep the function signature, behavior, and test results identical to the reference solution. Initialize variables used by the algorithm; do not introduce unused variables or redundant assignments merely to satisfy the style. Preserve data dependencies and keep initializations that must occur within loops or branches in their original context.\\
\textbf{Anchor extraction.} Select the leading initialization block together with the first subsequent main-logic statement as a single anchor span. Preserve their order and indentation.
\end{quote}
\subparagraph{GSM8K / MATH500 — math reasoning anchor.}
\begin{quote}\small
\textbf{Style.} Rewrite the solution as a short, simple chain of arithmetic steps, with at most one operation per step. Avoid algebraic manipulation that requires more than one variable at a time, and avoid abbreviations.\\
\textbf{Rewrite rule.} Preserve the reference numerical answer. Each step should be directly verifiable from the previous one.\\
\textbf{Anchor extraction.} Pick the single step that is most important to the final answer as the anchor, excluding the final answer itself.
\end{quote}

\subparagraph{HumanEval / MBPP — code reasoning anchor.}
\begin{quote}\small
\textbf{Style.} Rewrite the reference solution using only basic Python control flow (\texttt{for}, \texttt{while}, \texttt{if}, \texttt{return}) and elementary data structures (\texttt{list}, \texttt{dict}, \texttt{set}, \texttt{tuple}). Avoid advanced language features such as list/dict/set comprehensions, generator expressions, lambdas, decorators, or one-liners that fuse multiple operations.\\
\textbf{Rewrite rule.} Keep the function signature and behavior identical to the reference solution and pass all unit tests.\\
\textbf{Anchor extraction.} Pick the single line that most directly captures the core algorithmic step (e.g., the key conditional, accumulator update, or result-construction line) as the anchor.
\end{quote}

\subsection{Metric Details}
\label{appendix: metric}

We provide details for the four controllability metrics summarized in Section~\ref{sec:setup}. For each metric, the resulting control score is then combined with the corresponding task performance via the geometric mean to form the composite controllability score reported in the main paper.

\paragraph{Commented code (coding).}
This metric measures how densely and meaningfully comments are interleaved with code, rather than dumped as a leading block. After applying the same code post-processing used for $\mathrm{pass@1}$, we tag each line as either \texttt{code}, \texttt{comment}. After collapsing consecutive identical tags, the control score is the fraction of adjacent positions at which the tag switches between \texttt{code} and \texttt{comment}. A higher value indicates better interleaving.

\paragraph{Structured code (coding).}
We evaluate structural compliance using the abstract syntax tree (AST) of the target function.
After removing the function docstring and skipping permitted leading early-exit guards, we inspect the ordered statements directly within the function body.
An initialization is an assignment or annotated assignment to a single variable whose value is a supported literal or a zero-argument built-in constructor, such as \texttt{x = 0}, \texttt{res = []}, or \texttt{seen = set()}.
Assignments nested within loops, conditionals, or helper functions do not count toward the leading initialization block.

A solution passes if the remaining function body begins with at least one qualifying initialization and the contiguous initialization block is followed by a non-initialization statement other than a docstring or \texttt{pass}.
A single initialization is sufficient, and additional initializations receive no extra credit.
Let $s_i \in \{0,1\}$ indicate whether the $i$-th solution satisfies this constraint.
The control score is
\[
    S_{\mathrm{str}} = \frac{1}{N}\sum_{i=1}^{N} s_i,
\]
where $N$ is the total number of generated solutions.
Empty or unparseable outputs receive a score of zero.
This metric checks structural compliance; it does not assess whether the initialized values are semantically necessary.

\paragraph{Backward reasoning (math).}
This metric is designed for GSM8K and detects whether a solution is written in a goal-first, backward-reasoning style (e.g., ``To find $X$, we need $Y$\,''), as opposed to standard forward chain-of-thought (``First..., Then..., Finally...''). After isolating the current answer block, we apply two complementary gates and count a sample as backward only if it passes both:
\begin{itemize}
    \item \textbf{Rule gate ($G_1$)}: a keyword-based check that fires when the response contains backward cues such as ``to find $\ldots$ we need $\ldots$'' or ``working backwards $\ldots$''.
    \item \textbf{LLM-judge gate ($G_2$)}: an LLM judge (Qwen3-14B with $3$-seed majority vote) that verifies the semantic reasoning logic genuinely starts from the goal and traces back to the required premises, rather than relying on superficial cue words.
\end{itemize}
The control score is the intersection pass rate, i.e., the fraction of samples for which $G_1 \!=\! 1$ and $G_2 \!=\! 1$. Combining the two gates avoids both shallow keyword matching and overly lenient semantic judgments.

\paragraph{Formal math (math).}
This metric is also designed for GSM8K and rewards solutions written as symbolic, equation-driven derivations (e.g., \texttt{Let $x_1 = \ldots$}, \texttt{Let $x_2 = \ldots$}) rather than natural-language explanations. After the same cleaning and answer isolation as above, each non-empty line is classified into one of: \texttt{let}, \texttt{assign}, \texttt{eqonly}, \texttt{answer}, \texttt{eqprose}, \texttt{prose}, or \texttt{noise}. The control score is the fraction of formal lines:
\begin{equation}
F_1 \;=\; 100 \cdot \mathrm{mean}\!\left(\frac{\#\texttt{let} + \#\texttt{assign} + \#\texttt{eqonly} + \#\texttt{answer}}{\#\text{total lines}}\right).
\end{equation}

\subsection{Anchor Content Ratio}
\label{app:content_ratio}

Table~\ref{tab:content_ratio} reports the average anchor
length and the average fraction of output tokens
contributed by anchors on each benchmark.

\begin{table}[t]
\caption{Average anchor length and content ratio
across benchmarks.}
\label{tab:content_ratio}
\centering
\small
\setlength{\tabcolsep}{6pt}
\begin{tabular}{@{}lrrrr@{}}
\toprule
& HumanEval & MBPP & GSM8K & MATH500 \\
\midrule
Anchor length (tokens)
& 11.08 & 11.45 & 27.85 & 26.63 \\
Anchor content ratio (\%)
& 4.33 & 4.47 & 10.88 & 10.86 \\
\bottomrule
\end{tabular}
\end{table}
\subsection{Generalization across DLMs}
\label{app:generalization}

Table~\ref{tab:generalization} reports results on the source
and target DLMs using the anchor generator trained with
LLaDA-8B-Base outputs, without further training on
LLaDA-8B-Instruct.

\begin{table}[t]
\caption{Transfer from LLaDA-8B-Base to LLaDA-8B-Instruct
without retraining the anchor generator.
Formal, Backw., and Cmt. report controllability measured
by G-mean; Acc. and Pass@1 report task performance.
Parentheses indicate absolute gains over each DLM's
vanilla baseline.}
\label{tab:generalization}
\centering
\small
\setlength{\tabcolsep}{3pt}
\resizebox{\linewidth}{!}{%
\begin{tabular}{ll|ccc|cc|cc}
\toprule
\multirow{2}{*}{DLM}
& \multirow{2}{*}{Method}
& \multicolumn{3}{c|}{GSM8K}
& \multicolumn{2}{c|}{HumanEval}
& \multicolumn{2}{c}{MBPP} \\
& & Formal & Backw. & Acc.
& Cmt. & Pass@1 & Cmt. & Pass@1 \\
\midrule
\multirow{2}{*}{LLaDA-8B-Base}
& Vanilla
& 5.01 & 5.01 & 66.03
& 14.59 & 34.15 & 29.77 & 39.80 \\
& AnchorDLM
& \textbf{69.33} (+64.32)
& \textbf{71.08} (+66.07)
& \textbf{67.17} (+1.14)
& \textbf{43.89} (+29.30)
& \textbf{37.20} (+3.05)
& \textbf{40.05} (+10.28)
& \textbf{43.60} (+3.80) \\
\midrule
\multirow{2}{*}{LLaDA-8B-Instruct}
& Vanilla
& 17.08 & 30.94 & 56.10
& 29.83 & 39.02 & 28.25 & 40.80 \\
& AnchorDLM (transferred)
& \textbf{59.51} (+42.43)
& \textbf{62.34} (+31.40)
& \textbf{58.23} (+2.13)
& \textbf{45.02} (+15.19)
& \textbf{39.63} (+0.61)
& \textbf{47.80} (+19.55)
& \textbf{41.00} (+0.20) \\
\bottomrule
\end{tabular}%
}
\end{table}
\label{app:generalization_llada15}

\end{document}